%% file: root.tex
\documentclass[letterpaper, 10 pt, conference]{ieeeconf}  

\IEEEoverridecommandlockouts                              

\usepackage{graphicx} 
\usepackage{amsmath} 
\usepackage{amssymb}  
\usepackage{xcolor}
\usepackage{multirow}
\usepackage{makecell}
\usepackage{algorithm}
\usepackage{algpseudocode}
\makeatletter
\let\NAT@parse\undefined
\makeatother
\usepackage{hyperref}

\title{\LARGE \bf
SHIRE: Enhancing \underline{S}ample Efficiency using \underline{H}uman \underline{I}ntuition in \underline{RE}inforcement Learning
}

\author{Amogh Joshi, Adarsh Kosta, and Kaushik Roy\\
Purdue University, West Lafayette, IN 47907, USA\\$\{joshi157, akosta, kaushik\}@purdue.edu$
}

\begin{document}

\maketitle
\thispagestyle{empty}
\pagestyle{empty}

\begin{abstract}
The ability of neural networks to perform robotic perception and control tasks such as depth and optical flow estimation, simultaneous localization and mapping (SLAM), and automatic control has led to their widespread adoption in recent years. Deep Reinforcement Learning (DeepRL) has been used extensively in these settings, as it does not have the unsustainable training costs associated with supervised learning.
However, DeepRL suffers from poor sample efficiency, i.e., it requires a large number of environmental interactions to converge to an acceptable solution. Modern RL algorithms such as Deep Q Learning and Soft Actor-Critic attempt to remedy this shortcoming but can not provide the explainability required in applications such as autonomous robotics. Humans intuitively understand the long-time-horizon sequential tasks common in robotics. Properly using such intuition can make RL policies more explainable while enhancing their sample efficiency. In this work, we propose SHIRE, a novel framework for encoding human intuition using Probabilistic Graphical Models (PGMs) and using it in the Deep RL training pipeline to enhance sample efficiency. Our framework achieves $25-78\%$ sample efficiency gains across the environments we evaluate at negligible overhead cost. Additionally, by teaching RL agents the encoded elementary behavior, SHIRE enhances policy explainability. A real-world demonstration further highlights the efficacy of policies trained using our framework.


\end{abstract}

\section{Introduction}
\input{sections/introduction}

\section{Related Work}
\input{sections/related_work}

\section{SHIRE Framework}
\input{sections/shire_framework}

\section{Experiments}
\input{sections/experiments}



\section{Conclusion}
\input{sections/conclusion}

\textbf{Acknowledgement:}
This work was supported by 
CoCoSys, one of the seven centers in JUMP 2.0, an SRC program sponsored by DARPA



\bibliographystyle{IEEEtran}
\bibliography{root}

\end{document}

%% file: sections/introduction.tex
In recent years, AI has been applied to almost all aspects of human life. In particular, machine learning techniques have found applications in diverse areas such as fraud detection, medical diagnosis, image and pattern recognition, and language translation. Machine learning has also been of great interest in the robotic-vision community, where it has been successfully applied to perception tasks such as depth \cite{Zhu_2019_CVPR, junayed2022himodehybridmonocularomnidirectional, shen2022panoformerpanoramatransformerindoor} and optical flow estimation \cite{Gehrig3dv2021, lee2022fusion, mint, flowformer}, semantic segmentation \cite{Kirillov_2023_ICCV, Kirillov_2019_CVPR, segformer}, Simultaneous Localization and Mapping (SLAM) \cite{deepfactors}, and automatic control \cite{joshi2024realtimeneuromorphicnavigationintegrating}.

Among the prevalent machine learning techniques, supervised learning has been the most widely adopted, primarily due to its simplicity of implementation. While supervised learning is simple to implement, its simplicity belies its true cost in terms of data and compute requirements. The training cost of modern supervised learning models is of the order of hundreds of ZFLOPs ($1$ $Zettaflop=10^{21}FLOPs$) \cite{epoch2024trackinglargescaleaimodels}, with some state-of-the-art models requiring 21 YFLOPs ($1$ $Yottaflop=10^{24}FLOPs$) to train \cite{epoch2024trackinglargescaleaimodels}. Supervised training requires large, well-labeled and well-balanced data sets that are notoriously difficult to generate \cite{gebru2021datasheets}, especially for robotic control tasks. In addition, training time and cost are directly proportional to dataset size. Thus, larger datasets lead to even larger training overheads, making supervised training infeasible for autonomous robotics applications.

Another major shortcoming of existing machine learning approaches is their ``black-box" nature. Machine learning models tend to be neither explainable nor interpretable \cite{ribeiro2016should}. This problem is especially acute in supervised learning models, where a lack of clear decision rules and the absence of an explicit reward signal make it impossible to trace the model's ``reasoning" for arriving at a certain decision. This makes supervised learning untenable in safety-critical applications such as autonomous robotics, where both of these qualities are required.

Reinforcement Learning (RL) is a machine learning technique that trains an agent to perform a task through repeated interaction with the environment along with a system of rewards and punishments. A major advantage of RL is that while training requires a large number of interactions with the environment, these are still computationally cheaper than training on large supervised datasets. This makes RL ideally suited for tasks such as autonomous robotics, where datasets are either too small or do not exist at all due to data generation difficulties. Recently, deep reinforcement learning, where a neural network is used as the agent, has been successfully applied to autonomous robotics tasks such as dextrous in-hand manipulation \cite{handmanip} and quadrupedal walking \cite{haarnojalearning}.
Despite these successes, Deep RL has few drawbacks. Convergence to a good policy is dependent on factors such as random initialization, the quality of the reward function, and the representation granularity of the environment. Another shortcoming is the large amount of interactions required to converge to a good policy. Such large data requirement may result in poor sample efficiency in Deep RL algorithms.
These shortcomings have traditionally been remedied by supplying the agent with a model of its environment (also called its world). This approach is called Model-Based Reinforcement Learning (MBRL). While MBRL improves sample efficiency significantly, it is subject to instability issues caused by biases in the world model. Previous works such as \cite{psrl, cpsrl} have shown that exploiting prior knowledge of an environment's transition dynamics can lead to provably better sample efficiency in Deep RL settings \cite{Hafner2020Dream, Kaiser2020Model, janner2019trust}. However, such environmental information may not always be available in the real world.

Humans possess an intuitive understanding of the solutions to many robotics tasks. Though not always optimal, such intuition contains some component of the optimal solution for the task \cite{myers2010intuition}. Additionally, the causal reasoning displayed by humans shows an innate understanding of the long-term effects of current actions \cite{myers2010intuition}. Harnessing this intuition and causal reasoning in RL training can enhance sample efficiency and make the learned policy more explainable. However, encoding human intuition in a compute-friendly manner is still an open problem.

We propose SHIRE (Enhancing \underline{S}ample Efficiency using \underline{H}uman \underline{I}ntuition in \underline{RE}inforcement Learning), a framework for formalizing the process of encoding human intuition and using this formalism in conjunction with standard Deep RL algorithms to enhance sample efficiency. Policies trained on SHIRE are ``explainable" in the sense that they learn the elementary behavior encoded in the SHIRE framework. To the best of our knowledge, this framework is the first of its kind. The primary contributions of our work are summarised below:
\begin{itemize}
    \item A framework (SHIRE) for encoding human intuition as a Probabilistic Graphical Model (PGM) and using it to enhance the sample efficiency and explainability of Deep RL.
    \item Provide experimental evidence to show that our framework improves performance by $25\%$ for simple environments and more than $78\%$ for complex environments.
    \item Show significant reduction in training time and required samples for multiple tasks.
    \item Integrate a policy trained using our framework with a real robot and provide a video demonstration of the same.
\end{itemize}


%% file: sections/related_work.tex
To the best of our knowledge, no works targeted at enhancing sample efficiency use human intuition to add an inductive bias to training samples from an agent's rollout/experience buffer. In the following section, we provide a broad overview of the research directions prevalent in current literature aimed at enhancing sample efficiency and explainability in Reinforcement Learning.

Early efforts to improve sample efficiency in RL focused on model-based approaches.
Recent works include "Dream to Control" by Hafner et al.\cite{Hafner2020Dream}, which uses imagined trajectories to learn environment models. The authors in \cite{Kaiser2020Model} use the SimPLe algorithm combined with video prediction for policy learning in low-data settings. Janner et al. \cite{janner2019trust} branch off from real off-policy data to generate short world-model-generated rollouts for policy training. While these methods show promising results, they face challenges related to accurately modeling environments, bias in model-generated data, and computational overhead that scales with model complexity.

The challenges of world modeling led to the development of alternative approaches that focus on increasing sample efficiency through algorithm design. Entropy maximization techniques for both on-policy \cite{o2016combining} and off-policy \cite{haarnoja2017reinforcement} algorithms improve exploration and robustness by augmenting the reward maximization objective with entropy maximization. Soft Actor-Critic (SAC) \cite{haarnoja2018soft} is a model-free, off-policy algorithm that combines the entropy maximization approach with an actor-critic formulation for continuous state and action spaces, enabling sample-efficient policy learning. However, SAC requires a large experience replay buffer, leading to high memory overhead, and suffers from instability due to inaccurate posterior sampling by the actor network.

The policies discussed so far lack explainability, which is crucial in real-world robotics. Explainability in robotics can be interpreted as the ability to explain a policy's decisions based on (a) the environment it interacts with or (b) the agent's understanding of its own dynamics. PILCO \cite{deisenroth2011pilco} relies on the first interpretation, and addresses environmental factors using a Gaussian process model but underestimates future uncertainty by ignoring temporal correlations. DeepPILCO \cite{gal2016improving} improves upon PILCO with Bayesian Deep models but relies on access to the reward function and world state, quantities usually unavailable in practice. Chua et al. \cite{chua2018deep} enhance future-state uncertainty estimates using an ensemble of probabilistic and deterministic world models, achieving asymptotic performance close to that of model-free methods with the PETS algorithm. However, they still suffer from error accumulation over long task horizons.

Mutti et al.'s \cite{cpsrl} C-PSRL algorithm is similar to SHIRE in using causal graph priors to inform the learning process, but they serve different purposes. C-PSRL uses causal priors to learn a full causal model of the environment, which can be transformed into a Factorized MDP (FMDP) for more efficient RL, as shown by \cite{chen2020efficient, osband2014near}. However, exact planning in FMDP models is computationally intractable \cite{mundhenk2000complexity, lusena2001nonapproximability}. Additionally, since the causal priors used in C-PSRL encode prior knowledge of the environment's dynamics, they do not impart desirable behavior to the agent, leading to a lack of explainability. A comprehensive analysis is further hampered by the lack of experiments in this work. In contrast, SHIRE uses causal graphs, called "Intuition Nets," to model human knowledge of the agent (not the environment), making it model-free, computationally simpler, and more explainable, addressing C-PSRL's main shortcomings.

%% file: sections/shire_framework.tex
In this section, we provide a brief overview of Markov Decision Processes, standard Deep-RL training algorithms, and how our framework integrates with them.

\subsection{Markov Decision Processes}
Reinforcement Learning problems are defined as a finite-episode Markov Decision Process (MDP). An MDP ($\mathcal{M}$) consists of a tuple $(\mathcal{S}, \mathcal{A}, p, r, T)$, where $\mathcal{S}$ is the state-space of the environment with cardinality $S$, $\mathcal{A}$ is the action space of the agent with cardinality $A$, $p$ is a Markovian transition model such that $p(s_{t+1}|s_t, a)$ is the conditional probability of the next state $s_{t+1}$, given the current state $s_t$ and action $a$. $r:\mathcal{S}\times\mathcal{A}\rightarrow[0,1]$ is the deterministic reward function over a state-action pair $(s,a)$. $T$ denotes the episode horizon.

In every episode, an agent interacts with the environment in the following manner: an initial state is randomly chosen from an initial state distribution in $\Delta(\mathcal{S})$, where $\Delta(\mathcal{S})$ is the probability simplex over $\mathcal{S}$. At every step $t<T$, the agent selects an action $a_t\in\mathcal{A}$, and receives a reward $r_t\sim r(s_t, a_t)$. At the same time, the state transitions from $s_t$ to $s_{t+1}\sim p(\cdot|s_t, a_t)$. The episode ends either when $t=T$ or when the task is complete.

The strategy employed by the agent while choosing what action $a_t$ to take in state $s_t$ is defined by a stochastic policy $\{\pi_t\}_{t\in T}\in\Pi$, where $\Pi$ is the policy space and each $\pi_t$ is a function such that the conditional probability of choosing action $a$ in state $s$ at step $t$ is given by $\pi_t(a|s)$. To evaluate the quality of a policy $\pi_t$, we use the value function $V_t^\pi(s):\mathcal{S}\rightarrow[0, T]$ which is the expected sum of rewards received using $\pi$ when starting in a state $s$ and step $t$. Therefore, the value function can be defined as shown in Eq. \ref{valfunc}
\begin{equation}
    V_t^{\pi}(s):=\mathop{\mathbb{E}}_\pi\Bigg[\sum_{t'=t}^Tr(s_{t'}, a_{t'})\bigg|s_t=s\Bigg] \forall s\in\mathcal{S}, t\in[T]
    \label{valfunc}
\end{equation}

The optimal policy $\pi^*$ has the value function $V^*_t(s)$ given by
\begin{equation}
    V^*_t(s) = \mathop{argmax}_\pi V_t^{\pi}(s)
\end{equation}

Policy optimization algorithms in Deep RL collect experience through an agent's interaction with the environment and use it to attempt to converge to a policy $\pi'$ whose value function $V_t^{\pi'}(s)$ is as close as possible to $V^*_t(s)$. 
Due to the first-order optimizations, small learning rates, stochastic policies, and inefficient exploration employed by almost every policy optimization algorithm, these methods suffer from poor sample efficiency.

\subsection{Policy Optimization Algorithm}
There exist a plethora of RL policy optimization algorithms in literature. Of these, the most well-known are PPO \cite{ppo}, DQN \cite{dqn}, and SAC \cite{haarnoja2018soft}. In this work, we train our policies using a modified version of the PPO algorithm. The PPO algorithm's loss function is as shown in Eq. \ref{eq:ppo_loss}
\begin{equation}
    loss_{PPO} = loss_{policy} + loss_{entropy} + loss_{value}
    \label{eq:ppo_loss}
\end{equation}
where $loss_{policy}$ is the policy loss computed using the clipped advantage function, $loss_{entropy}$ is the degree of uncertainty in the policy, and $loss_{value}$ is the mean squared error between the expected and actual reward. The SHIRE framework adds an additional term to this loss, which is described in the next section.

\begin{figure}[ht]
    \centering
    \includegraphics[scale=0.58]{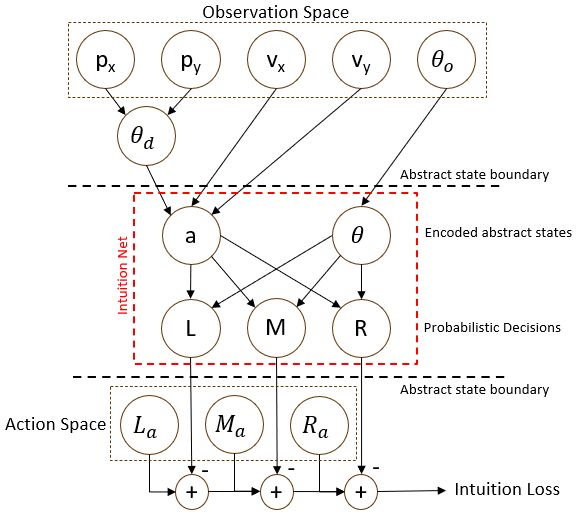}
    \caption{Intuition Encoding for the Lunar Lander Environment}
    \label{fig:int_enc}
\end{figure}

\subsection{Intuition Encoding}
Humans possess a remarkable ability to develop an intuitive understanding of any problem they solve \cite{myers2010intuition}. This ability is particularly apparent in the long time horizon, sequential tasks common in robotics, where each action must be taken after considering its effects over the entire task horizon. While not always the optimum solution for a task, such human intuition contains the necessary basic understanding of either the task or the robot's dynamics needed to learn the task efficiently. Such intuition has the potential to significantly accelerate RL training, thus enhancing sample efficiency, and adding an implicit notion of explainability to the policy. Encoding human intuition in a machine-friendly manner is, however, a non-trivial task.

Probabilistic Graphical Models (PGMs) encode the dependence between random variables as structured graphs. Bayesian Networks are a special class of PGMs that encode the joint distribution of a set of random variables using Directed Acyclic Graphs (DAGs). As demonstrated in literature \cite{koller2009probabilistic}, Bayes Nets can capture dependence (or independence) between random variables in a compact, compute-friendly, factorized form. Another advantage of Bayes Nets is their ability to encode epistemic uncertainty implicitly in the form of probability distributions. For these reasons, SHIRE uses Bayes Nets to encode simple task-specific human intuition.

\begin{figure*}[ht]
    \centering
    \includegraphics[width=\textwidth]{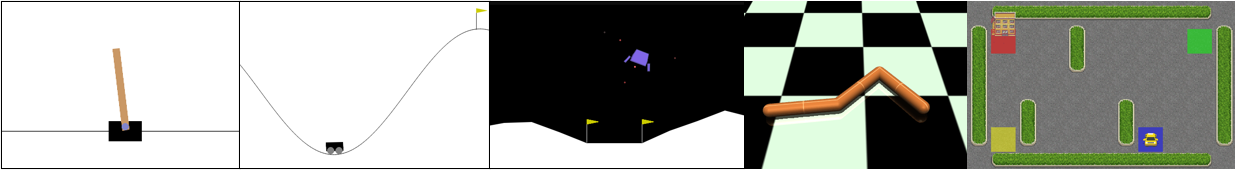}
    \caption{Gymnasium environments used for evaluating the SHIRE framework. From left to right: CartPole, MountainCar, LunarLander, Swimmer, and Taxi.}
    \label{fig:rl_envs}
\end{figure*}

The SHIRE framework can be broadly divided into three stages - Intuition Net construction, Abstract State encoding, and Intuition Loss computation. Fig. \ref{fig:int_enc} diagrammatically illustrates this workflow for the Lunar Lander environment \cite{lunarlander}. The goal of this environment (shown in Fig. \ref{fig:rl_envs}) is to land the craft as smoothly as possible at the landing site designated by the two flags using three thrusters - the main, and the left and right orientation thrusters. The agent's observation consists of the $x$- and $y$- components of its position and velocity, as well as its orientation and angular velocity. The intuition for this environment is simple - the velocity vector of the craft must be directed at the landing site at all times. The agent is expected to learn to aim the bottom of the craft at the landing zone by itself. Note that other intuitions are also possible. The following paragraphs explain how the SHIRE framework uses this intuition to speed up the training of an RL policy for the LunarLander task.

\vspace{1mm}
\begin{algorithm*}[ht]
    \begin{algorithmic}
        \caption{Intuition Loss Computation}\label{alg:int_loss_comp}
        \Require $batch \sim Rollout Buffer$ \Comment{sample batch of size n from the rollout buffer}
        \State $a \gets A(batch)$ \Comment{Get actions from rollout buffer}
        \State $o \gets O(batch)$ \Comment{Get observations from rollout buffer}
        \For{$i\gets 1, n$} \Comment{n is the batch size}
            \State $s_i \gets S(o)$ \Comment{encode abstract states}
            \State $e_i \gets PGM(s_i)$ \Comment{compute ``intuitive" actions using probabilistic inference}
            \State $m_i \gets M(e_i, a_i)$ \Comment{compute mismatch vector}
        \EndFor
        \State $loss_{intuition} \gets \sum_{i=1}^n max(0, 1 - m_ia_i)$ \Comment{convert mismatch vector to hinge loss}
        \State \Return $loss_{intuition}$ \Comment{return intuition loss}
    \end{algorithmic}
\end{algorithm*}
\vspace{-1mm}
\subsubsection{Intuition Net Construction}
We call the task-specific PGMs encoding human intuition ``Intuition Nets". Since the intuition networks are used to compute a loss term, the child nodes of the intuition net always correspond to the action space of the task. In the case of the Lunar Lander environment, this yields three child nodes as shown in Fig. \ref{fig:int_enc}, each node corresponding to one of the three thrusters - Main ($M$), Left Orientation thruster ($L$), and Right Orientation thruster ($R$). Each of the three nodes has two states corresponding to the ``$idle$" and ``$fire$" states of the thrusters. Controlling the craft's velocity vector requires knowledge of the craft's orientation and the direction of desired acceleration. Therefore, the parent nodes of the intuition net are the desired acceleration direction ($a$) and the craft orientation ($\theta$). Note that node $a$ has only two states $positive$ and $negative$ corresponding to the direction of desired acceleration, while node $\theta$ has four states - one for each of the four quadrants. Due to the simplicity of Intuition Net, and to ensure that the desired intuition is accurately reflected, we choose to hand-code the probability assignments governing inference in this intuition net. We intentionally introduce uncertainty into the intuition net to prevent the policy from memorizing the encoded intuition.

\subsubsection{Abstract State Encoding}
In this stage, observation values from the recorded rollout buffer are encoded into the abstract states of the PGM's parent nodes. Since the landing site is located at the origin, the observed values of the craft's position yield the desired orientation of the velocity vector ($\theta_d$) as $\theta_d = \tan^{-1}\frac{p_y}{p_x}$
This desired orientation, combined with the observed values of the velocity components $v_x$ and $v_y$ yields the desired acceleration direction, which is encoded as the abstract state of the node $a$. The observed value of the craft's orientation is thresholded to yield the abstract state of the node $\theta$.

\subsubsection{Intuition Loss Computation}
In this stage, the actions generated by the policy are compared against the corresponding predictions of the Intuition Net, and mismatches between the two are stored in a mismatch vector. However, the sum of such a vector isn't convex. Therefore, to ensure convergence, we convert it into a convex hinge loss termed the ``Intuition Loss". This procedure is described in Algorithm \ref{alg:int_loss_comp}. In our implementation, the for loop in Algorithm \ref{alg:int_loss_comp} is vectorized for fast computation.
This intuition loss is computed for every batch and is then added to the loss, as shown in Eq. \ref{eq:totalloss}. For Lunar Lander, we choose to penalize mismatches in the state of the main thruster more than mismatches in the orientation thrusters due to the larger effect the main thruster has on the craft's state. 
\begin{equation}
    loss = loss_{PPO} + loss_{intuition}
    \label{eq:totalloss}
\end{equation}
\begin{figure}[ht]
    \centering
    \includegraphics[scale=0.7]{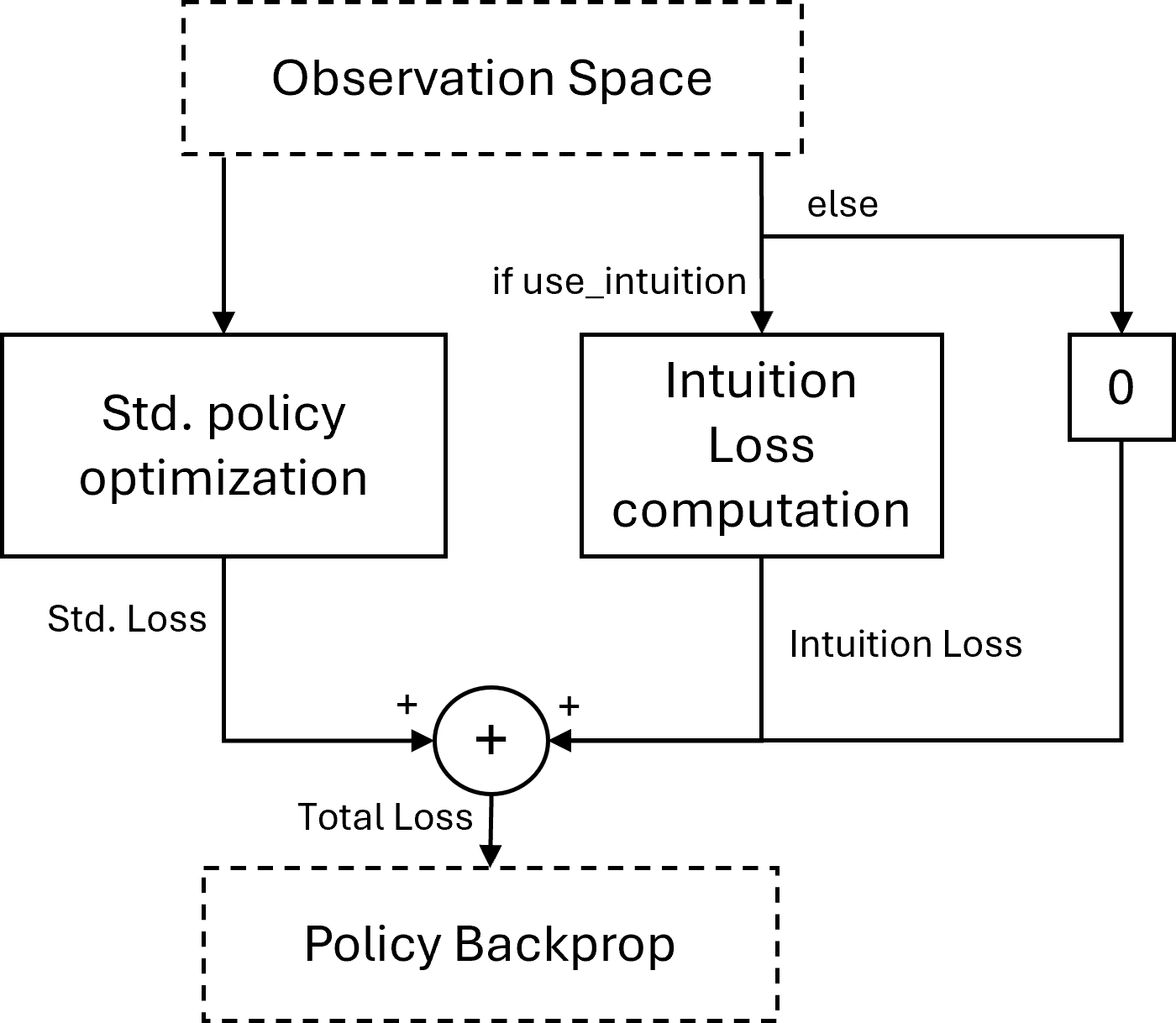}
    \caption{Integration of Intuition Loss with existing RL Policy Optimization Algorithms}
    \label{fig:int_loss_flow}
\end{figure}
Since the abstract state encoding input to the Intuition Net depends only on observation data from the rollout buffer, it is agnostic to the policy optimization algorithm being used. Thus, the SHIRE framework can be integrated into any existing policy optimization algorithm by simply adding a flag argument, as shown in Fig. \ref{fig:int_loss_flow}. In the next section, we present results for the SHIRE framework on a variety of environments using Proximal Policy Optimization (PPO), an on-policy algorithm.

%% file: sections/experiments.tex
To assess the efficacy of the SHIRE framework, we conduct policy optimization experiments on various standard gym environments. Across environments, we tuned a two-layer vanilla agent with 64 neurons per layer and stored the best result. We then enabled our framework and re-ran the best vanilla experiment on each environment with the same seed for comparison. Thus, the vanilla and SHIRE experiments differ only in the intuition loss scaling coefficient.
Table \ref{tab:results} shows the number of samples (i.e., interactions with the environment) required to attain a solved state both with and without SHIRE. In this section, we describe the intuition encoded in each Intuition Net and provide quantitative analyses of the improvements achieved by using our framework and of the overheads imposed by this framework. Note that we consider an environment to be ``solved" when a policy achieves an average reward greater than or equal to the ``Solved State Reward" (SSR) for that environment over 100 consecutive evaluation episodes.

\subsection{Encoded Intuitions}
For each of the tested environments, ``$f$" is the intuition loss computation function, and the intuition encoded by the Intuition Net, as well as the definition of a ``solved state" is as given below:
\subsubsection{Cart Pole Environment}
The Cart Pole environment, where an agent attempts to keep an inverted pendulum upright, is the simplest environment for which we provide results. A policy is said to have ``solved" this environment when it obtains an average reward of $500.0$ over 100 episodes. Intuitively, if the pendulum isn't upright, one moves the cart towards the pendulum's lean. This intuition is encoded as a simple two-node Intuition Net in SHIRE, yielding an intuition loss given by Eq. \ref{eq:cartpole}, where $\theta$ is the angle made by the pole with the vertical.
\begin{equation}
    intuition\hspace{1mm}loss = f(\theta)
    \label{eq:cartpole}
\end{equation}

\subsubsection{Mountain Car Environment}
Mountain Car is a deterministic MDP first introduced by Andrew  Moore \cite{moore1990efficient} in 1990. The goal of this environment is to apply force on the car to propel it to the top of the mountain as shown in Fig. \ref{fig:rl_envs}. This environment defines ``solving" as obtaining an average reward greater than $-110.0$ over 100 episodes. Since climbing the hill in one go is impossible, the logical solution is to oscillate up and down opposite sides of the mountain until enough momentum has been built up to climb it. This intuition can be further simplified to say ``Apply force in the direction of the car's current velocity". Again, SHIRE encodes this simplified intuition as a two-node Intuition Net given by Eq. \ref{eq:mountaincar}, where $v$ is the velocity of the car.
\begin{equation}
    intuition\hspace{1mm}loss = f(v)
    \label{eq:mountaincar}
\end{equation}

\subsubsection{Lunar Lander Environment}
The Lunar Lander environment is based on the Atari game of the same name where the player is tasked with landing a craft as smoothly as possible within a landing zone designated by two flags. The only actions an agent can take are independently firing (or not firing) the main and the left and right orientation thrusters. An agent that achieves an average reward greater than $200.0$ over 100 episodes is said to have ``solved" this environment. A naive intuition for solving this environment is ensuring that the landing craft's velocity vector always points towards the landing zone, as explained in the previous section. This yields an intuition loss given by Eq. \ref{eq:lunarlander},
\begin{equation}
    intuition\hspace{1mm}loss = f(p_x, p_y, v_x, v_y, \theta_o)
    \label{eq:lunarlander}
\end{equation}
where $p_x, p_y$ are the $x$- and $y$-components of the lander's position, $v_x, v_y$ are the $x$- and $y$-components of the lander's velocity, and $\theta_o$ is the lander's orientation. Additionally, enforcing anti-parallelism between the velocity and orientation vectors of the landing craft ensures that the main thruster can be used for braking, resulting in a softer, more controlled landing. This in turn, further enhances sample efficiency, as shown in Table \ref{tab:results}. These ideas are encoded as five-node Intuition Nets, as shown in Fig. \ref{fig:int_enc}.

\begin{table*}[ht]
    \centering
    \vspace{2mm}
    \caption{Number of steps (interactions) required to solve standard gym environments. The Sample Efficiency gains are specified in cells adjacent to the corresponding parameter. SSR: Solved State Reward, BBR: Best Baseline Reward.}
    \vspace{-2mm}
    \begin{tabular}{|c|c|c|c|c|c|c|c|}
    \hline
        \multirow{2}{*}{Environment} & \multirow{2}{*}{SSR/BBR} & \multicolumn{2}{c|}{PPO Baseline} & \multicolumn{4}{c|}{PPO with SHIRE} \\
        \cline{3-8}
        & & \makecell{N(steps)\\to solve} & \makecell{Time to solve\\(minutes)} & \makecell{N(steps)\\to solve} & Gain (\%) & \makecell{Time to solve\\(minutes)} & Gain (\%) \\
        \hline
        CartPole & 500.0 & 8192 & 2.07 & 5120 & \textbf{37.5} & 2.51 & \textbf{-21.25} \\
        MountainCar & -110.0 & 510k & 150.96 & 110k & \textbf{78.43} & 36.29 & \textbf{75.96} \\
        LunarLander (w/o anti-parallelism) & 200.0 & 120k & 10.81 & 90k & \textbf{25} & 6.83 & \textbf{36.82} \\
        LunarLander (w anti-parallelism) & 200.0 & 120k & 10.81 & 70k & \textbf{41.67} & 5.3 & \textbf{50.97} \\
        Swimmer & 110.0 & 3.37M & 804 & 1.395M & \textbf{58.61} & 313.8 & \textbf{60.97} \\
        Taxi & 8.1 & 1.19M & 104.34 & 845k & \textbf{28.99} & 72 & \textbf{30.99} \\
        \hline
    \end{tabular}
    \label{tab:results}
\end{table*}

\subsubsection{Swimmer Environment}
The Swimmer environment is a variable-length, multi-segment robot control environment introduced by Remi Coulum \cite{coulom2002reinforcement}. The goal of this environment is for the agent to make the robot (called a ``swimmer") move in the positive $X$-direction as much as possible. The only action that the agent may take to accomplish this task is to apply torques on the hinge joints between adjacent segments. The larger number of independent state and control variables in this environment makes it significantly more challenging than the other environments presented thus far. However, despite the greater difficulty, the intuition for solving this environment is simple - the torque applied to adjacent joints must be opposed in direction. This helps enforce the serpentine motion necessary for solving this task. The intuition loss for this environment is given by Eq. \ref{eq:swimmer}
\begin{equation}
    intuition\hspace{1mm}loss = f(\theta_1, \theta_2)
    \label{eq:swimmer}
\end{equation}
where $\theta_i$ is the angle between the links of the $i^{th}$ joint.

Swimmer is an ``unsolved" environment, i.e., no set SSR value exists. Therefore, to ensure a fair comparison, we train a policy without our framework and use the best reward achieved by it as a proxy for the SSR. We call this reward the ``Best Baseline Reward" (BBR). Sample Efficiency is calculated as the number of interactions (steps) taken to reach this reward value. Our experiments use a three-element robot.

\subsubsection{Taxi Environment}
The Taxi environment is a grid-world navigation task where the agent (a taxi driver) must navigate to one of four fixed locations to pick up a passenger and then transport them to their destination at another of the four points. At every step, the agent can move one square up, down, left, or right.
The grid world contains obstacles of unknown size at unknown locations, adding to the complexity of finding the necessary paths. This environment encodes the taxi's position, the passenger's initial position, and his destination into a single integer, yielding an intuition loss as in Eq. \ref{eq:taxi} where $o$ is the encoded observation.
\begin{equation}
    intuition\hspace{1mm}loss = f(o)
    \label{eq:taxi}
\end{equation}
Taxi, like Swimmer, is an unsolved environment. Therefore, to evaluate the sample efficiency gains achieved by our framework, we use the same strategy as we used for the Swimmer environment. We implement this environment in the real world using a TurtleBot and an NVIDIA Jetson Nano, a video demo of which is attached to this manuscript.

\begin{table}[ht]
    \centering
    \caption{SHIRE Computational Overheads ($\mu s$/sample)}
    \begin{tabular}{|c|c|c|}
    \hline
        Environment & \makecell{Intuition\\Net Size} & \makecell{Overhead per Sample\\($\mu s$)} \\
        \hline
        CartPole & 2 & 223 \\
        MountainCar & 2 & 215 \\
        \makecell{LunarLander\\(w/o antiparallelism)} & 5 & 254\\
        \makecell{LunarLander\\(w/ antiparallelism)} & 6 & 257\\
        Swimmer & 4 & 232\\
        Taxi & 4 & 235\\
        \hline
    \end{tabular}
    \label{tab:overheads}
\end{table}

\subsection{Sample Efficiency Gains}
The results shown in Table \ref{tab:results} highlight the fact that even trivially simple intuition can make a significant difference in RL performance. SHIRE achieves $>25\%$ sample efficiency gains across all tested environments.
In CartPole, the simplest environment we provide results for, we achieve a $37.5\%$ sample efficiency gain at the cost of $21.25\%$ more wall clock time. This worsening of training time can be explained by the simplicity of the CartPole environment, which allows for very fast simulation, causing worse wall clock time performance, even for overheads as small as the one shown in Table \ref{tab:overheads}. In the MountainCar environment, SHIRE pushes the agent to learn the oscillatory behavior necessary for the solution, yielding gains of 78\% and 76\% in sample efficiency and wall-clock time respectively. In Lunar Lander, the naive velocity intuition yields a modest 25\% sample efficiency gain. Enforcing anti-parallelism between the velocity and craft orientation vectors boosts this gain to 41\%! Teaching a Swimmer agent the necessary serpentine motion leads to sample efficiency gains of more than  58\% while using observation data such as the taxi's location and the passenger's destination to encourage the agent to move toward the target position improves sample efficiency by 29\%. We observe that performance gains increase proportional to the complexity of the environment. This makes intuitive sense, as a better understanding of the basics (encoded into our intuition nets) aids the learning of complex tasks. This behavior also proves our hypothesis that SHIRE makes policies explainable by teaching them elementary behavior. The improvements in sample efficiency are accompanied by a small training-time overhead associated with the intuition loss computation, which is explained in the following section.

\subsection{SHIRE Overheads}
The latency overheads incurred by the Intuition Loss computation pipeline are approximately proportional to the size of the Intuition Net. Table \ref{tab:overheads} shows the intuition loss computation overhead for all the tested environments. It is immediately apparent that the overhead imposed by SHIRE is negligibly small compared to the average time per environment step, which is typically of the order of tens of milliseconds. Any increase in training time caused by this overhead is compensated for by SHIRE's sample efficiency gains, as shown by the gains in wall-clock time shown in Table \ref{tab:results}.

%% file: sections/conclusion.tex
We present SHIRE, a novel, compute-friendly framework for encoding task-specific human intuition as a PGM to enhance RL training sample efficiency and explainability. Experiments show SHIRE provides over 25\% gains in sample efficiency across environments, with minimal computational overhead, which is offset by the improved efficiency.

This work enables rapid prototyping of RL policies, facilitating efficacy checking for various policy architectures, thereby improving the RL policy development life cycle. We hope this work spurs further research in efficient and explainable RL, enabling the development of robust RL policies for safety-critical tasks such as autonomous flight.